%% file: main.tex
 \documentclass[sigconf]{acmart}
\AtBeginDocument{%
  \providecommand\BibTeX{{%
    \normalfont B\kern-0.5em{\scshape i\kern-0.25em b}\kern-0.8em\TeX}}}

\DeclareMathOperator*{\argminA}{arg\,min} 
\DeclareMathOperator*{\argmaxA}{arg\,max} 

\newcommand\numberthis{\addtocounter{equation}{1}\tag{\theequation}}
\usepackage{balance}
\usepackage[english]{babel}

\copyrightyear{2021}
\acmYear{2021}
\setcopyright{acmcopyright}
\acmConference[MM '21]{Proceedings of the 29th ACM International Conference on Multimedia}{October 20--24, 2021}{Virtual Event, China}
\acmBooktitle{Proceedings of the 29th ACM International Conference on Multimedia (MM '21), October 20--24, 2021, Virtual Event, China}
\acmPrice{15.00}
\acmDOI{10.1145/3474085.3475525}
\acmISBN{978-1-4503-8651-7/21/10}

\acmSubmissionID{mfp1907}

\begin{document}
\fancyhead{}

\title{Diverse Multimedia Layout Generation with Multi Choice Learning}

\begin{teaserfigure}
    \includegraphics[width=\textwidth, scale=0.1]{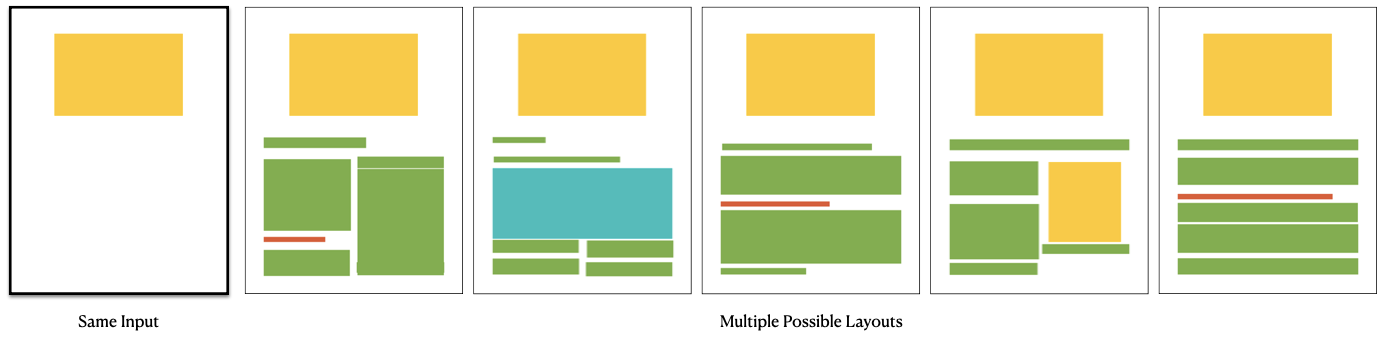}
    \caption{From the same input, LayoutMCL generates diverse and plausible document layouts with different composition of object types and sizes. LayoutMCL can also create multimedia layouts for mobile applications and magazines.}
    \Description{Layout Recommendation}
\end{teaserfigure}






\author{David D. Nguyen}
\email{d.d.nguyen@unsw.edu.au} 
\orcid{0000-0002-8243-195X}
\affiliation{
  \institution{UNSW Sydney}
  \country{Australia}
}
\additionalaffiliation{%
    \institution{CSIRO, Data61}
    }
\additionalaffiliation{%
    \institution{Cybersecurity CRC}
    }
    
\author{Surya Nepal}
\email{surya.nepal@data61.csiro.au} 
\orcid{0000-0002-3289-6599}
\affiliation{
  \institution{CSIRO, Data61}
  \country{Australia}
}
\authornotemark[2]
    
\author{Salil S. Kanhere}
\email{salil.kanhere@unsw.edu.au} 
\orcid{0000-0002-1835-3475}
\affiliation{
  \institution{UNSW Sydney}
  \country{Australia}
}
\authornotemark[2]

\renewcommand{\shortauthors}{Nguyen, et al.}

\input{src/0_abstract}



\maketitle

\section{Introduction}
\input{src/1_introduction}
\input{src/2_related}
\input{src/3_approach}
\input{src/4_evaluation}

\input{src/5_conclusion}
\input{src/6_acks}
\bibliographystyle{ACM-Reference-Format}
\balance 
\bibliography{layout}

\end{document}

%% file: src/0_abstract.tex
\begin{abstract}

Designing visually appealing layouts for multimedia documents containing text, graphs and images requires a form of creative intelligence. 
Modelling the generation of layouts has recently gained attention due to its importance in aesthetics and communication style.
In contrast to standard prediction tasks, there are a range of acceptable layouts which depend on user preferences. 
For example, a poster designer may prefer logos on the top-left while another prefers logos on the bottom-right.
Both are correct choices yet existing machine learning models treat layouts as a single choice prediction problem. 
In such situations, these models would simply average over all possible choices given the same input forming a degenerate sample.
In the above example, this would form an unacceptable layout with a logo in the centre.

In this paper, we present an auto-regressive neural network architecture, called \textbf{LayoutMCL}, that uses multi-choice prediction and winner-takes-all loss to effectively stabilise layout generation.
\textbf{LayoutMCL} avoids the averaging problem by using multiple predictors to learn a range of possible options for each layout object.
This enables \textbf{LayoutMCL} to generate multiple and diverse layouts from a single input which is in contrast with existing approaches which yield similar layouts with minor variations. 
Through quantitative benchmarks on real data (magazine, document and mobile app layouts), we demonstrate that LayoutMCL reduces Fréchet Inception Distance (FID) by 83-98\% and generates significantly more diversity in comparison to existing approaches.

\end{abstract}


%% file: src/1_introduction.tex
Layout design consists of the spatial arrangement of heterogeneous objects in graphical multimedia such as documents, presentations or website designs. 
This creative task is generally human-led and requires an understanding of relationships between objects (e.g. text, images) to communicate an aesthetic style and message \cite{shipman1995finding}. 
The resurgence of neural networks has led to projects that seek to mimic human creative intelligence by learning patterns in visually appealing designs \cite{nguyen2015innovation,elgammal2017can}.
Recent interest in modelling layout designs stems from opportunities to build superior recommendations in graphical editing applications \cite{li2019layoutgan} or automated graphical honeypots \cite{bowen2009baiting}.
Generative layout models also have the potential to be applied in procedural content generation for games \cite{sorenson2010towards}, reinforcement learning environments \cite{justesen2018illuminating} and virtual/augmented reality design \cite{beever2020leveled}.


Layout modelling involves the prediction of a set of object bounding boxes, categories and their interdependent relationships.
For any layout in the training data, each prediction can be considered conditional on both the prior partial layout and the creator's personal preferences. 
A partial layout is observable and provides supporting evidence for alignment and potential spatial arrangements.  
However, creator preferences are only observable to the creator and cannot be easily serialized into training data.
The inability to capture this information in the training data leads to an unobserved stochastic process. 

\begin{figure}[h]
  \centering
  \includegraphics[width=200pt]{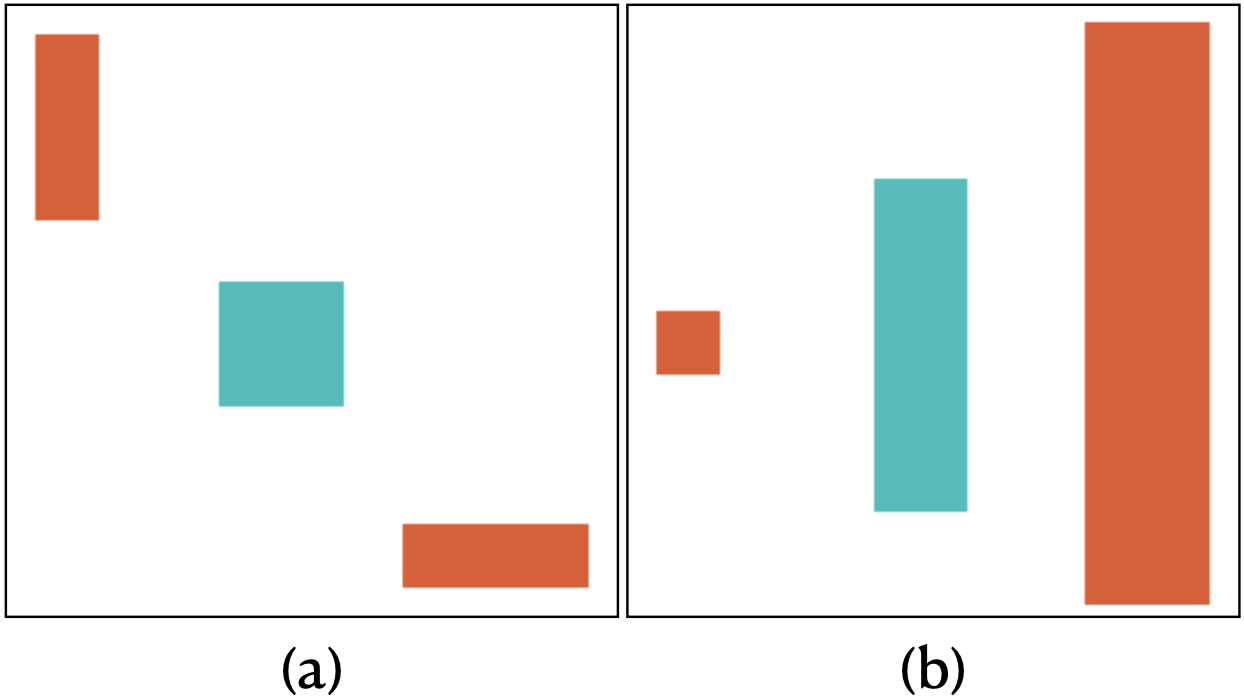}
  \caption{This visualises the "Averaging Problem" caused in 2 scenarios with one input value and more than one ground truth (red). In both scenarios, a single neural network minimises the mean-squared error between two equally likely ground truths given the same input values. The network simply takes the average of both ground truths across 4 dimensions (x,y,w,h), resulting in a prediction (green) that satisfies neither label.}
  \Description{The Averaging Problem}
   \label{fig::avgproblem}
\end{figure}

Furthermore, creator preferences are all equally correct yet may lie on different extremes of the output distribution. 
For example, a poster designer may prefer logos on the top-left while another designer prefers them on the bottom-right.
These preferences imply that layouts are a multi-output problem, $f(x) = \{y_1, y_2 ... y_n\}$, where there are more than one acceptable outputs given the same input. 
However, existing approaches \cite{li2019layoutgan,patil2019read,zheng2019content,lee2020neural} treat layout recommendation as a single choice prediction problem. 
An architecture with a single regression predictor would take the mean of all possible outputs for a given input \cite{rodriguez2018action,firman2018diversenet}.
Using the above example, this would result in a medium sized icon located in the middle of the page which is demonstrated in 2 scenarios in Figure~\ref{fig::avgproblem}.
We refer to this as the \textbf{averaging problem} which results in degenerate layouts that do not satisfy any ground truth and has been observed in prior work \cite{zheng2019content,lee2020neural,patil2019read,li2019layoutgan}. 
Any feasible model must structure layouts as a multi-output problem to manage the un-observable stochastic processes in the training data.

From the opposite perspective, users of the generative neural network would also have their own stylistic preferences for layouts. 
Recent work in \cite{lee2020neural} has used the concept of \textit{user constraints} via graphs to address this issue.
User constraints are captured as prespecified relationships between objects including location and sizes.
For example, the user specifies that an object must be "larger than" another object via an edge relationship. 
This approach is referred to as conditional layout generation, which is in contrast to our unconditional generation method.

Another approach presented in \cite{zheng2019content} uses generative adversarial networks (GANs) \cite{goodfellow2014generative} and \textit{content awareness}.
Content Awareness  refers to user preferences with specified keywords such as "bride, party, dance". 
Developing a large dataset with these keywords is a very time consuming process and difficult to scale for other solutions. 
Moreover, these approaches can only propose a single final layout given a constraint or partial layout.
This implies that all users have relatively the same preferences which is highly unlikely.

In this paper, we present a multi-choice learning architecture called \textbf{LayoutMCL} that generates diverse multimedia layouts to represent the range of user preferences.
Based on auto-regressive convolutional and recurrent networks, our approach diverges from existing approaches by framing layout generation as a sequence of multiple choices. 
At each step, LayoutMCL predicts a new object category, bounding box and whether to stop generation based on the partial layout. 

In contrast to existing approaches, LayoutMCL has the capability to optimize for multiple outputs within different distributions by using multiple predictors.
This is achieved in a single architecture by combining a mixture layer with Winner-Takes-All (WTA) loss \cite{guzman2012multiple}.
Our approach boosts likelihood of accurate hypotheses and reduces the likelihood of unused predictors.
Using these techniques, LayoutMCL generates diverse layout arrangements and is capable of generating multiple plausible layouts from a single input as shown in Figure 1.
This flexibility allows users to choose a final layout from a wide selection based on their own preferences. 

To evaluate performance, we use layout datasets from a wide variety of graphical multimedia domains including mobile apps \cite{deka2017rico}, documents \cite{zhong2019publaynet} and magazines \cite{zheng2019content}.
These benchmarks include quantitative measurements where we demonstrate that LayoutMCL outperforms existing models by reducing Frichet Inception Distance (FID) by 83-98\% on all datasets and Alignment by 16-35\% for Mobile Apps and Documents. 
FID measures similarity by quantifying the distance between generated and real layout. 
The Alignment metric measures the level of alignment between object bounding boxes within each layout.

This paper is organized as follow. 
In Section 2, we review existing work on generative models for layouts and multi-output learning algorithms.
In Section 3, we provide an overview of the LayoutMCL architecture 
and then delve into the mechanics of our proposed multi choice learning framework for layouts. 
In Section 4, we conduct several experiments to demonstrate the strengths of LayoutMCL in comparison to existing layout generative models. 
These experiments include quantitative benchmarks and a diversity and consistency test.
We end with an interesting demonstration of LayoutMCL providing multiple layout recommendations in a hypothetical graphical editor.


%% file: src/2_related.tex
\section{Related Work}
\subsection{Layout Generation}
One of the earliest approaches by Li et al. in \cite{li2019layoutgan}, utilizes GANs to re-arrange graphical layouts objects using wire-frame renderings. 
Another GAN approach by the authors of \cite{zheng2019content} introduces contextual awareness using keywords focusing mainly in the domain of magazine layouts. 
While a feasible approach, pairing keywords with layouts can be a time consuming process which limits its transferability to other datasets.

A framework called LayoutVAE \cite{jyothi2019layoutvae} uses variational auto-encoders (VAE) \cite{kingma2013auto} to generate stochastic scene layouts with a two stage process: categorical count and bounding box prediction. 
Patil et al. in \cite{patil2019read} adapt a combined recursive network and VAE framework by \cite{li2019grains} for document layouts using only several hundred training examples.
The authors in \cite{lee2020neural} use a GNN framework based on \cite{johnson2018image} to generate complete graphs based on partial graphs using size and relative layout location. 

These approaches treat bounding box placement as a single point estimate or uni-modal distribution and thus remains vulnerable to degenerate samples as reported in \cite{zheng2019content,lee2020neural,patil2019read,li2019layoutgan}.
In contrast, our approach frames layouts as a sequence of multiple choices enabling predictions in different distributions. 

\subsection{Multi Choice Learning}
To address the problem of using single point estimates in problems with different outputs, multi-choice learning has recently emerged as a viable tool.
The term multi-choice learning can be traced to \cite{guzman2012multiple,guzman2014multi} which introduced a structured multi-hypotheses SVM algorithm using winner-takes-all loss (WTA). 
In a two stage training regime, a set of predictors generate a set of hypotheses and only the predictor which produced the minimum loss is updated during training. 
Using this technique, \cite{kirillov2015inferring} showed that increasing diversity can be achieved by penalizing similar outputs. 
WTA was developed for neural network ensembles in \cite{lee2016stochastic} which encourages each network to become an "expert". 
This boosted output diversity and was applied for a range of tasks: image classification, semantic segmentation, image captioning and image synthesis \cite{chen2017photographic}.

A variant called Relaxed WTA (RWTA) appeared in \cite{rupprecht2017learning} which interestingly showed that multi-hypothesis prediction (MHP) resulted in Voronoi tessellation in the output space.
RWTA applies a small update to all non-minimum predictors during training. 
This relaxation helps move outputs away from an initial single Voronoi cell and therefore increases diversity.
MHP motivated further work with RNN multi-choice prediction in \cite{bhattacharyya2018accurate}.
A multi label approach was developed in \cite{firman2018diversenet} that uses a control vector to induce multiple outputs.
These methods by \cite{rupprecht2017learning,firman2018diversenet} demonstrate that multi-choice learning can be harnessed by a single neural network architecture, instead of an ensemble seen in \cite{lee2016stochastic}. 
This simplified training into a single regime and reduced training times.

\subsection{Multimodal Learning}
An alternate approach to multi-output problems is representing different peaks of $y$ with the same input $x$ as separate Gaussian distributions. 
This is advantageous over multi-choice frameworks because it provides an estimation of uncertainty through the variance.
Mixture density networks (MDN) introduced in \cite{bishop1994mixture} generates mixture Gaussian distributions from the outputs of neural networks.
Interestingly, MDNs use a mixture coefficient layer that forms a probability density function over each mixture distributions.
These were applied in real-valued domains such as generating speech and handwriting \cite{graves2013generating}.
MDNs at higher dimensions are difficult to optimize with gradient descent due to numerical instability, mode collapse and require special initialization schemes \cite{rupprecht2017learning,cui2019multimodal,makansi2019overcoming,zhou2020movement,prokudin2018deep}.

A recent sample and fit framework by \cite{makansi2019overcoming} showed that MDNs can be stabilized for future prediction tasks such as pedestrian and car movement.
This framework combined concepts from multi-choice learning by developing Evolving WTA Loss (EWTA) for its sampling phase. 
In a successive round of updates, the EWTA updates the top $k$ hypotheses and decreases $k$ at each step until $k=1$. 
This reduces hypotheses being stuck in an unacceptable regions between ground truths as seen in RWTA.
Another variant, Entropy WTA was introduced by \cite{zhou2020movement} to prevent modal collapse for MDNs in robotic primitive movement tasks.
Modal collapse occurs when a sub-set of mixtures are not trained due to lack of data. 
During generation, these under-trained mixtures are selected and result in degenerate samples.
Despite these improvements to MDNs, we still experience numerical instabilities for layouts caused by divisions of very small variances resulting in large gradients.

Our approach combines several ideas from multi-choice learning and mixture density networks. 
Multi-hypothesis prediction is more stable and easier to train than MDNs however is vulnerable to collapse when hypotheses exceed labels. 
This is due to the difficulty in determining which predictor has not been trained for which input.
As a result, we use the idea of mixture coefficient layers from MDNs   
to help resolve this problem by diminishing the likelihood choosing untrained predictors and vice versa.
This new framework will be discussed at further depth in Section~\ref{sec::mcf}. 






%% file: src/3_approach.tex
\section{LayoutMCL}

Section~\ref{sec::arch}  begins with a conceptual discussion that explains the intuition behind our overall layout architecture followed by a technical overview. 
Section~\ref{sec::mcf} delves into the mechanics of our multi choice learning framework.

\subsection{Architecture} \label{sec::arch}

Our architecture mimics how humans would design graphical layouts for a presentation or poster. 
If you were to observe the creative process in a time-lapse, from an empty canvas humans will place a series of objects in sequential order. 
Each decision is an mental interaction between other existing objects and a human's personal preference.
This is also how we construct speech and writing.
We already have a vague mental image of what we want to say or design before-hand which explains why some approaches apply GANs \cite{li2019layoutgan}.
However what matters most is that the human output process is always structured sequentially, and not in a single step.

\begin{figure}[h]
  \centering
  \includegraphics[width=\linewidth]{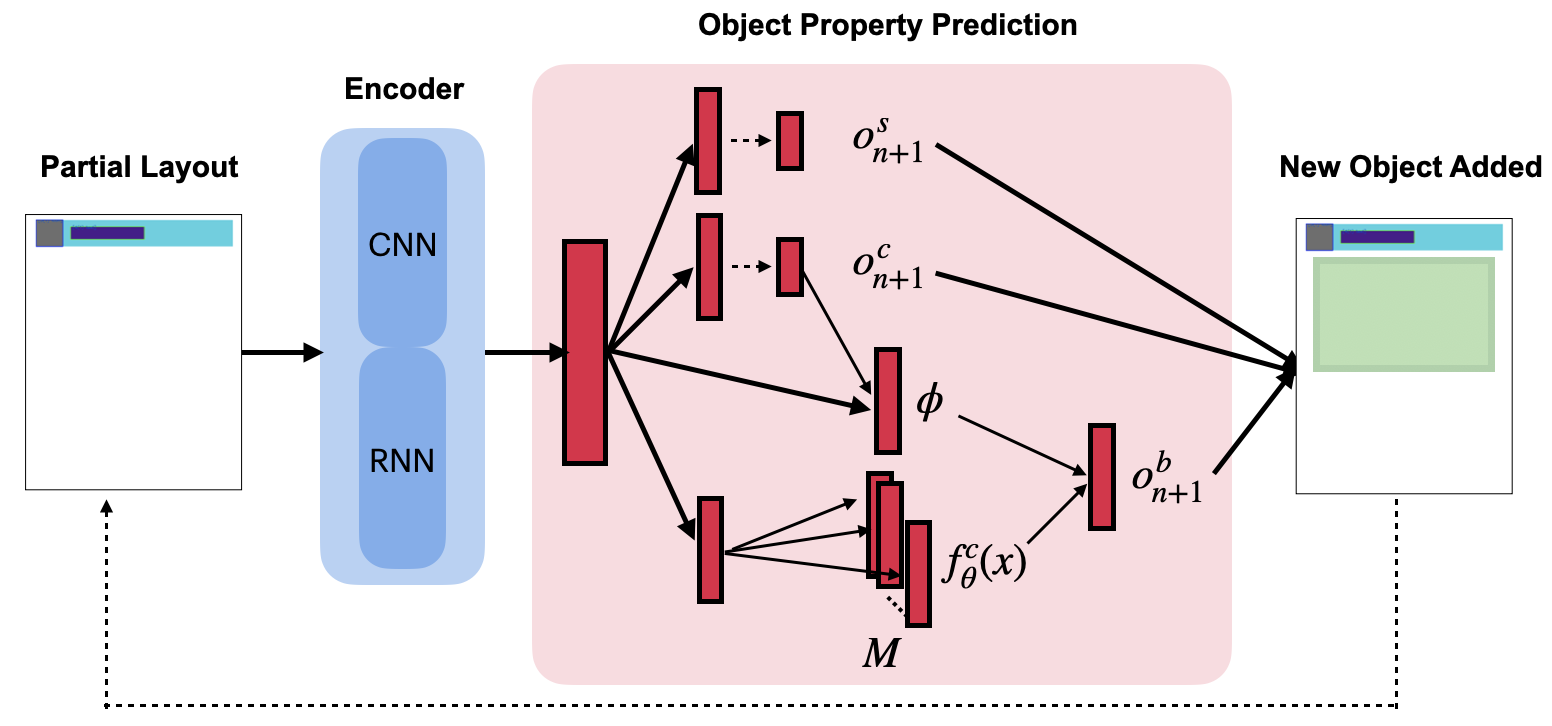}
  \caption{Auto-regressive architecture of LayoutMCL depicting a single step in object prediction from a partial layout to a new object.
  New objects are appended to the existing layout and generation continues until $o^s=1$}
  \Description{LayoutMCL}
   \label{fig::mclarch}
\end{figure}

Based on these intuitions, we design LayoutMCL to generate layouts using an auto-regressive architecture, shown in Figure~\ref{fig::mclarch}.
The input to LayoutMCL is the existing layout which is appended with a new object at the following step.  
More formally, a layout is comprised of a set of objects, $L_\pi = \{o_0, ..., o_n\}$, where $\pi$ is a specific object ordering.
Our experiments find that \textit{human-reading} order of the layout objects generates the best quality samples, particularly in multi-column documents. 

Each step consists of three object predictions created by separate prediction modules. 
Separate modules allows the outputs to take on different distributions types. 
The object bounding box, $o^b$, coordinates $(x,y,w,h)$ are normalized between $\{0,1\}$. 
The object category, $o^c$, varies by dataset (E.g. toolbar, title, icon).
The stop label, $o^s$ is 0 for all objects except for the final object where $o^s=1$. 
An object is therefore defined as $o_{n} = \{o_n^b, o_n^c, o_n^s\}$.


\subsubsection{CRN Encoder} \label{sec::encoder}
The encoder processes an existing layout using both visual and object categorical information in the same manner that a human would.
This is achieved by using  a joint convolutional neural network (CNN) and bidirectional recurrent network (RNN) which creates a shared representation. 
The RNN encodes the continuous geometric dimensions and categorical properties of $L_\pi$ to also avoid the problem of occlusion.
Bidirectionality is utilized because an object's location in $\pi$ is not solely based on prior objects.
This module aggregates the resulting vector to create $X_{layout}$.
The CNN is leveraged to distinguish visual patterns including alignment and spacing between locally spaced objects. 
This module encodes a rendered masked layout to build a \textit{spatial representation}, $X_{spatial}$. 
$X_{spatial}$ and $X_{layout}$ are combined into a latent vector, $X_{shared}$, shared between three prediction modules.

\subsubsection{Object Prediction Modules\label{sec::mhd}}
This bounding box module uses a multi-choice framework, discussed in the Section~\ref{sec::mcf}, to learn the diverse range of human preferences in object placement and size with multiple  predictors.
Each predictor can be thought of as learning a separate preference.
This bounding box module consists of a mixture coefficient layer and $M \times C$ distinct 2-layer feed-forward layer where $C$ is the number of categories and $M$ is the number of hypotheses/predictors. 
The predictors are 2-layer feed-forward networks that use an intermediate ReLU layer and final Sigmoid layer that predicts $o_{n+1}^b$.

The categorical module, $F_c$, is a 2-layer feed-forward networks with intermediate ReLU activation and a final log softmax layer that predicts $o_{n+1}^c$. 
$F_c$ is trained by minimizing the negative log likelihood loss, $l_c$, between the output and category labels. 
Likewise, the stop module, $F_s$, uses the same 2-layer feed-forward networks and predicts $o_{n+1}^s$.
$F_s$ is trained by minimizing the binary cross entropy loss, $l_s$.
In summary, the total loss minimized during training is defined in Equation~\ref{eq:lossfn}.
\begin{equation}\label{eq:lossfn}
    \mathcal{L}_{total} = l_c\lambda_c + l_s\lambda_s + l_b\lambda_b
\end{equation}
where the $l_b$ is the loss for the bounding box module defined in Equation~ \ref{eq::metaloss3}. 
This will be discussed in the next section.
$\lambda$ are hyper-parameters used to re-weight the losses during training. 

\subsection{Multi Choice Framework for Layouts\label{sec::mcf}}

Here we describe a simple multi choice framework that combines recent ideas from \cite{lee2016stochastic,rupprecht2017learning, makansi2019overcoming,zhou2020movement,firman2018diversenet}.
To begin, we discuss the two most prevalent problems found in past literature using multiple choice frameworks. 
\begin{enumerate}
    \item \textbf{Modal Collapse}. A subset of predictors are not \textit{paired} with ground truths and generate degenerate hypotheses during test time. Pairing refers to being trained with reference to a ground truth. 
    \item \textbf{Averaging Problem}. All predictors are paired, however some or all are paired with multiple ground truths. This results in some or all hypotheses lying in an unacceptable central regions between ground truths. 
\end{enumerate} 

\begin{figure*}[h]
  \centering
  \includegraphics[width=\linewidth]{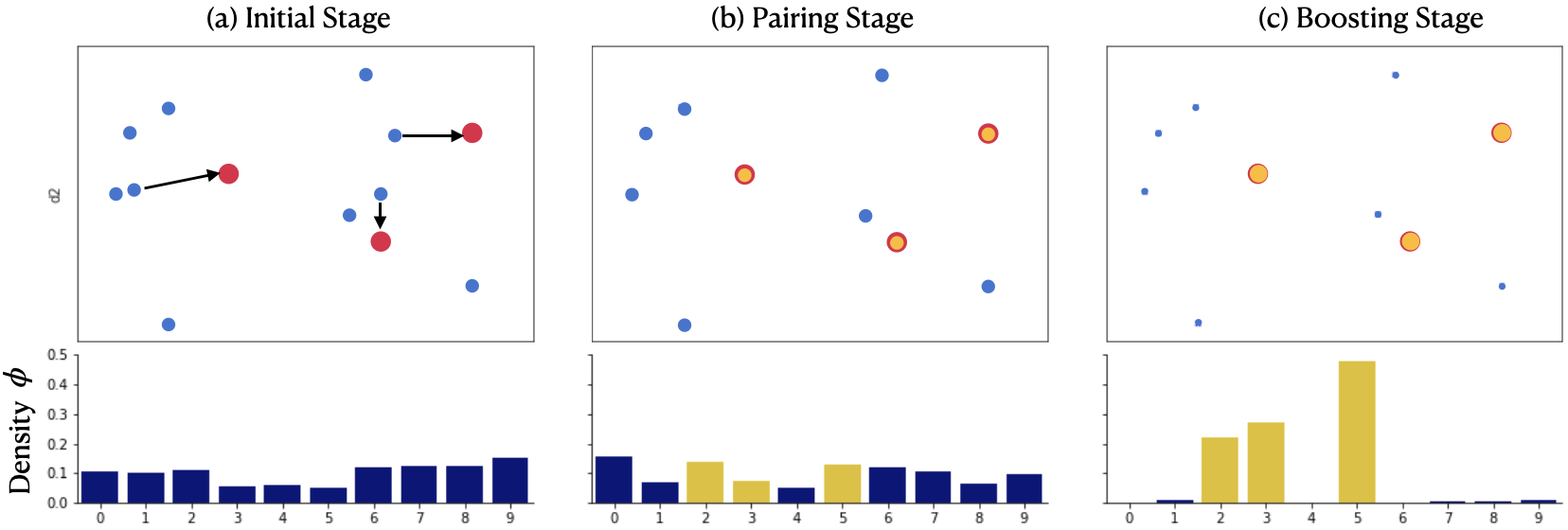}
  \caption{The three stages of our multi-choice framework depicting ground truths in red and hypotheses in blue. The top row shows movement of hypotheses over time. Bottom row shows change of density by predictor. (a) 10 predictors are randomly initialized with 3 ground truths. (b) Ground truth are paired with closest hypothesis (gold in bottom row) (c) Density of paired predictors are boosted and unpaired predictors pushed to zero.}
  \Description{Density and vanilla Loss}
   \label{fig::wtaours}
\end{figure*}

Balancing these two problems is not an easy task.
Early approaches \cite{lee2016stochastic,guzman2012multiple} focused on solving the averaging problem caused by single point estimates.
This was solved by using vanilla WTA loss however poorly initialized and unpaired predictors in regions far from ground truth labels are ignored leading to modal collapse. 
More recent approaches \cite{makansi2019overcoming,rupprecht2017learning} attempt to resolve modal collapse by moving all predictions towards ground truth labels with variants of WTA loss. 
A limitation of Relaxed WTA \cite{rupprecht2017learning} is the concentration of hypotheses in central regions between ground truth labels as seen in Figure~\ref{fig::wtaother}a.
Evolving WTA \cite{makansi2019overcoming} showed significant improvement in reducing the number of hypotheses in central regions, however some still remain as shown in Figure~\ref{fig::wtaother}b.

We take a different approach to this problem by combining a mixture coefficient layer with vanilla WTA loss.
The mixture coefficient layer predicts a probability density function over the set of predictors.
This idea circumvents modal collapse by boosting the coefficient of good predictors and reduces the coefficient of poor predictors.
This is useful in circumstances where the number of predictors exceed the number of ground truth labels.
In these situations, the likelihood of unpaired predictors are pushed towards zero and are unlikely to surface during test time.
In the opposite circumstance, all predictors are still paired with a ground truth.

Like \cite{rupprecht2017learning} we use multi-hypothesis prediction generated by multiple prediction layers, each initialized separately. 
However our framework is designed within an auto-regressive architecture and used for predicting objects with different categories that exhibit different output patterns.
For example, a title is generally smaller than an image. 
Each category, therefore, uses a separate sub-group of predictors to generate hypotheses.
In the below sections we describe our framework in detail. 



\subsubsection{Multi-Hypothesis Prediction}

Given a category, $c \in C$, and $x \in X_{shared}$, $f_\theta^c(x)$ is defined as a set of $M$ predictors parameterized by $\theta$:
\begin{equation}\label{eq:fc}
    f_\theta^c(x) = (f_{1}^c(x),...,f_{M}^c(x)).
\end{equation}
These predictors generates $M$ hypotheses where $\hat{y_i}$ refers to the $i^{th}$ hypothesis and $y$ is the corresponding ground truth label.
For the purpose of layouts, we use the L1 loss for the $i^{th}$ predictor.

\begin{equation}\label{eq:fc2}
    L_1(\hat{y_i}, y) = \left \rvert \lvert \hat{y_i} - y \right \rvert \rvert.
\end{equation}

A framework using a single hypothesis would simply take the mean over all ground truth labels leading to the averaging problem as discussed earlier.
Instead we first look at vanilla WTA loss \cite{makansi2019overcoming,rupprecht2017learning} defined in Equation~\ref{eq:metaloss1}.
\begin{align*}
    \mathcal{L}(f_\theta^c(x), y) &= \sum_{i=1}^M w_i L_1(\hat{y_i}, y) \numberthis \label{eq:metaloss1}  \\  
    w_j &= \delta(j = \argminA_i\left \rvert \lvert \hat{y_i} - y \right \rvert \rvert) \numberthis \label{eq:metaloss2}, 
\end{align*}
where $\hat{\delta}$ is the Kronecker delta returning 1 when true and 0 otherwise.
The Kronecker delta is used to select the best hypothesis $\hat{y_i} \in \hat{Y}$ that minimizes $L_1(\hat{y_i}, y)$.
In some sense, we are simply \textbf{pairing} predictors with ground truth labels.
Variations of WTA-loss alter the count of best hypotheses or the value of $\delta$ \cite{makansi2019overcoming,rupprecht2017learning}, however we take a different approach by introducing a mixture coefficient into this loss function.

\subsubsection{Mixture Coefficient Layer}

The use of a mixture coefficient layer means that we treat each predictor similar to a mixture in MDNs \cite{bishop1994mixture}.
This layer acts as a discrete probability function over predictors based on the shared representation and object category.
The mixture coefficient layer, $F_d: \mathcal{X} \to \mathcal{Y}$ is defined as:
\begin{equation}\label{eq:mixturelayer}
    \phi = F_d(x, c)
\end{equation}
where category is introduced as a concatenated one-hot vector and $\phi$ is a $M$ sized vector that sums to one. 
\begin{equation}\label{eq:mixtureequal}
    1 = \sum_{i=1}^M{\phi_i}.
\end{equation}
where $\phi_i$ is the coefficient of the $i^{th}$ predictor.
This vector is normalized with a softmax operation.

To combine this with Equation~\ref{eq:metaloss1}, the minimization of $\phi_i$ should increase the coefficients for the paired predictors and reduce the coefficients of unpaired predictors.
To achieve this, we minimize the negative log likelihood of $\phi$ as shown in Equation~\ref{eq::metaloss3}.

\begin{equation}\label{eq::metaloss3}
    \mathcal{L}(f_\theta^c(x), y) = \sum_{i=1}^M -log(\phi_i)w_i L_1(\hat{y_i}, y),
\end{equation}
During generation, predictors are sampled from a multinomial distribution parameterized by $\phi$.

\subsubsection{Stages}

The multi-choice framework can be demonstrated through a toy example shown in Figure~\ref{fig::wtaours}.
The top row of scatter plots show the movement of hypotheses (blue) over time in relation to the ground truths (red). 
The bottom row contains a bar chart showing the change of $\phi$ for each predictor.
There are three key stages that the multi-predictors and the mixture coefficient layers go through: 
\begin{enumerate}
    \item \textbf{Initial Stage} (Figure~\ref{fig::wtaours}a). 10 predictors and mixture coefficient layer with output vector, $\phi$, are randomly initialized along with 3 ground truths.
    The vector, $\phi$, shows that the density of all predictors are almost equal.
    \item \textbf{Pairing Stage} (Figure~\ref{fig::wtaours}b).
    Each ground truth is paired with the closest hypothesis.   The chosen hypothesis moves closer to the ground truth until their loss is negligible. 
    The vector, $\phi$, shows that the density of all predictors are almost equal, including the chosen (shown with gold bars) .
    \item \textbf{Boosting Stage} (Figure~\ref{fig::wtaours}c).
    Minimizing distance no longer decreases loss, therefore the network begins to boost $\phi_i$ for chosen predictors (gold) and decreasing $\phi_i$ for unpaired predictors. 
    The chance of sampling unpaired predictors is less than 99\% thus modal collapse becomes unlikely. 
\end{enumerate}
Despite appearances, it is worth noting that each stage does not require any initiation and begins automatically.

\subsubsection{Comparison to WTA Variants}

\begin{figure}[h]
  \centering
  \includegraphics[width=\linewidth]{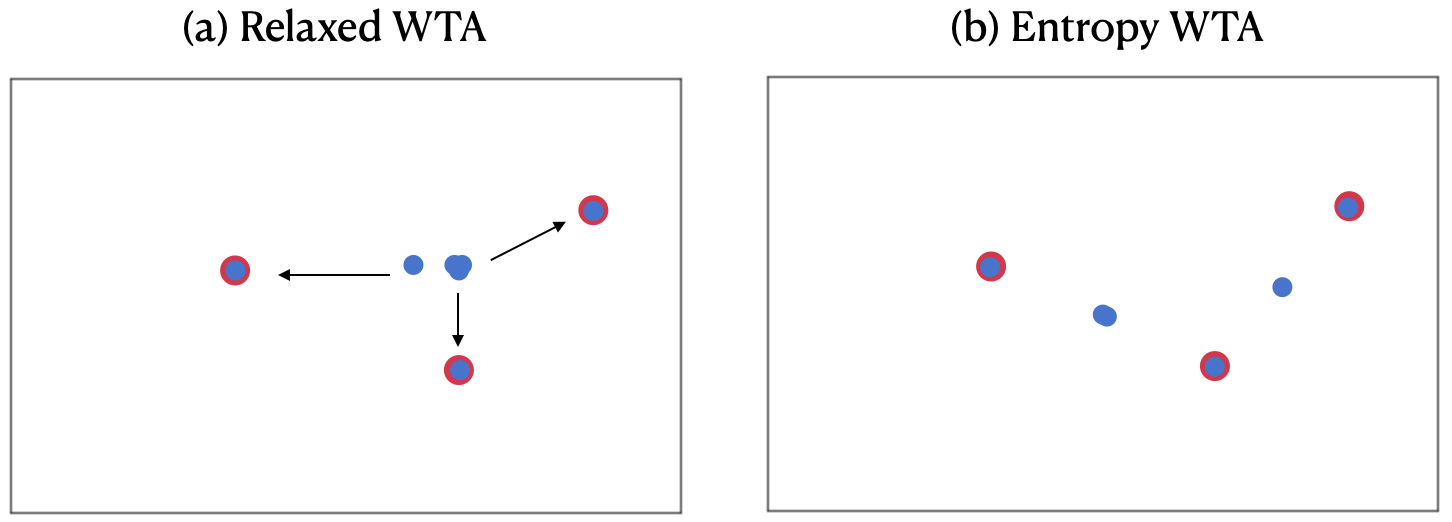}
  \caption{The final equilibrium for Relaxed and Evolving WTA.
  Hypotheses are shown with blue and ground truths with red circles. As seen in \cite{makansi2019overcoming}.}
  \Description{Other WTA Losses}
   \label{fig::wtaother}
\end{figure}

Using the exact same initialization scheme from the previous toy example, Figure~\ref{fig::wtaother} shows the final equilibrium of the 10 predictors using Relaxed and Evolving WTA. 
Figure~\ref{fig::wtaother}a shows that using Relaxed WTA results in large concentrations of hypotheses in central regions between ground truths. 
Figure~\ref{fig::wtaother}b shows that using Evolving WTA impressively moves hypotheses away from central regions towards ground truths.
On average we find that only 2-3 hypotheses remain stuck between regions of ground truths as seen in the figure. 
In comparison to Evolving WTA, the advantage of using a mixture coefficient layer for layouts is that the likelihood of sampling an unpaired hypothesis is very low. 
In the previous example, the probability of selecting an unpaired hypothesis is less than 1\%. 
In contrast, using Evolving WTA would raise the probability of selecting a poor hypothesis significantly to 20-30\%. 

The disadvantages of using the mixture coefficient layer, however, is that the likelihood of paired hypotheses are not equally likely. 
In Figure~\ref{fig::wtaours}c, we can see that predictor 5's density is significantly higher (49\%) than predictor 2 and 3. 
Since $\phi_i$ of unpaired hypotheses are pushed towards zero, we can rectify this problem during test time by recalculating $\phi_i$ for paired predictors as $\hat{\phi_i}=1/P$ where P is the number of paired predictors. 

\subsubsection{Number of Predictors}
A common question that naturally arises is \textit{"How many predictors should you use?"}.
The correct number depends on the dataset complexity and number of ground truths that arise during training for each input.
This is clearly difficult to calculate in multi-dimensional datasets due to exponential number of input-output combinations and has been a common criticism of multi-choice learning.

The use of a mixture coefficient layer allows us to offer some simple guidance based on our observations.
Our hypothetical architecture should initialize enough predictors so that the number of hypotheses sparsely cover all regions of the output space. 
In the worse case scenario, where ground truths form concentrated clumps in one dimension and are extremely spaced apart in another, coverage should be increased such that each ground truth has their own paired predictor. 
Over time, the mixture coefficient of non-paired predictors will diminish to zero, thus these should not surface.
Hence, the brief answer to the question is \textit{the more the better}, but this obviously has to be balanced with computation power and training speed.
Based on tests, a general rule of thumb is to start with 10 predictors and move up in multiples of 2.

%% file: src/4_evaluation.tex
\section{Evaluation}
\label{cha:evaluation}
In this section, we conduct experiments to demonstrate the strengths of LayoutMCL in comparison to existing layout generative models. 
These experiments include quantitative benchmarks, a diversity and consistency test. We also 
demonstrate LayoutMCL providing multiple recommendations in a hypothetical graphical editor.

\subsection{Experiment Setup}
\label{sec:setup}
This section provides implementation details of LayoutMCL and benchmark models.
We describe the datasets and propose metrics that measure the similarity between generated and real layouts. 

\paragraph{\textbf{Implementation}}
Our LayoutMCL is implemented in PyTorch and uses a bounding box module with 10 predictors per step.
The encoder consisting of 2 stacked bi-directional GRU layers with 128 hidden units and 5 CNN layers.
The network was trained with a learning rate of 0.001, batch size of 512 with an Adam optimizer.
The hyper-parameters in Equation~\ref{eq:lossfn} are $\lambda_c=1$,  $\lambda_s=1$  and $\lambda_b=40$.

\paragraph{\textbf{Benchmark Models}} 
Layout generative models are designed with different capabilities.
The closest two models to our approach are LayoutVAE \cite{jyothi2019layoutvae} and Neural Design Network \cite{lee2020neural}.
These models allow specification of object categories and do not require the specification of object counts and keyword pairings for all datasets.
\cite{patil2019read,li2019layoutgan,zheng2019content} do not satisfy all of these properties; therefore they are difficult to compare fairly.

\begin{enumerate}
    \item \textbf{LayoutVAE} \cite{jyothi2019layoutvae}. This model combines two conditional VAEs that operate in 2 successive stages: object count and bounding box. 
    In the first stage, the counts for each object are predicted using a Poisson distribution.
    In the second stage, each object's bounding boxes are generated.
    \item \textbf{Neural Design Network} \cite{lee2020neural}.
    NDN frames layouts as graphs where objects are the nodes and their respective location and sizes form relational edges. This model combines the use of graph convolutional networks \cite{johnson2018image} and VAEs to generate design layouts. 
    NDN-all is a format where \textbf{all} prespecified existing location and size relationships between objects are given prior to generation. 
    Since NDN-all performs best in \cite{lee2020neural}, this model is used for benchmarks in this paper.
\end{enumerate}

\paragraph{\textbf{Datasets}} Here we describe the three publicly available multimedia layout datasets from different domains. 
\begin{enumerate}
    \item \textbf{Mobile App} \cite{deka2017rico}. Following \cite{lee2020neural}, we only include portrait layouts with 13 categories (toolbars, images, text, icons, buttons, inputs, list items, advertisements, page indicator, web views, background images, drawers, modals) that have less than or equal to 10 objects. This leaves 21k examples.
    \item \textbf{Magazine} \cite{zheng2019content}. This dataset contains 4k layouts with 6 categories (text, images, headlines, over-image text, over-image headline, backgrounds)
    \item \textbf{Document} \cite{zhong2019publaynet}. Layouts are filtered for single and double column portrait documents with less than or equal to 10 objects per layout. The dataset uses all 5 categories (text, title, figure, table, list) which leaves 230k examples.
\end{enumerate}

\paragraph{\textbf{Metrics}}
The proposed metrics are designed to capture similarity of generated layouts to a held-out test dataset.
\begin{enumerate}
    \item \textbf{Alignment} \cite{lee2020neural}. This captures a common characteristic found in layouts; where objects tend to be left, centre or right aligned with other objects. 
    This generally improves the readability and aesthetics for the reader, particularly in magazines and documents. 
    This is defined in Equation~\ref{eq::alignment} as:
    \begin{equation} \label{eq::alignment}
        \frac{1}{N_d}\sum_d\sum_i\min_{j,i\neq j}\{\min(l(o^d_i,o^d_j),m(o^d_i,o^d_j),r(o^d_i,o^d_j)\}),
    \end{equation}
    where $N_d$ is count of generated layouts and $o^d_i$ is the $i^{th}$ object of the $d^{th}$ layout.
    Alignment is calculated using the normalized bounding box values.
    An alignment score closer to the test dataset is better.

\item \textbf{Fréchet Inception Distance (FID)} \cite{heusel2017gans}. 
FID was originally used to measure realism in GAN samples.
This has been adapted to measure the similarity between generated and real layouts. 
For each dataset, an encoder backbone (Section~\ref{sec::encoder}) is trained to discriminate between real and fake layouts. 
Fake layouts are created by taking existing layouts and adding perturbations drawn from the uniform distribution, $\mathcal{X} \sim U(-0.25,0.25)$, to each bounding box dimensions. 
The second-last output layer has 512 units and the FID distance between two samples is calculated with:
\begin{equation} \label{eq::fid}
    d^2 = ||mu_1 - mu_2||^2 + Tr(C_1 + C_2 - 2(C_1 \times C_2)^{1/2})
\end{equation}
where $mu_i$ is the output vector's mean, $C_i$ is the co-variance matrix and Tr is the trace operation.
The FID model's accuracy at detecting fake layouts from the entire held-out test dataset are: Mobile Apps (95.5\%), Magazines (91.2\%) and Documents (89.3\%).
We interpret a smaller FID score as better as it indicates higher similarity. 

\item \textbf{Fake Positive (\%)}. Using the same FID discriminator above, we compute Fake Positive as the percentage of samples the model assigns as fake over the total number of generated samples. 
This is defined as:
\begin{equation} \label{eq::fidfake}
    FP = \frac{1}{N}\sum^N_i{\delta(j = \argmaxA_i(\gamma_i))}
\end{equation}
where $\gamma$ is the 2D vector from the final discriminator layer and $j$ is the index corresponding to the fake category.
A model with a lower Fake Positive (\%) is better as it generates realistic layouts that "trick" the discriminator at higher rates.

\end{enumerate}

\begin{figure*}[h]
  \centering
  \includegraphics[width=\linewidth]{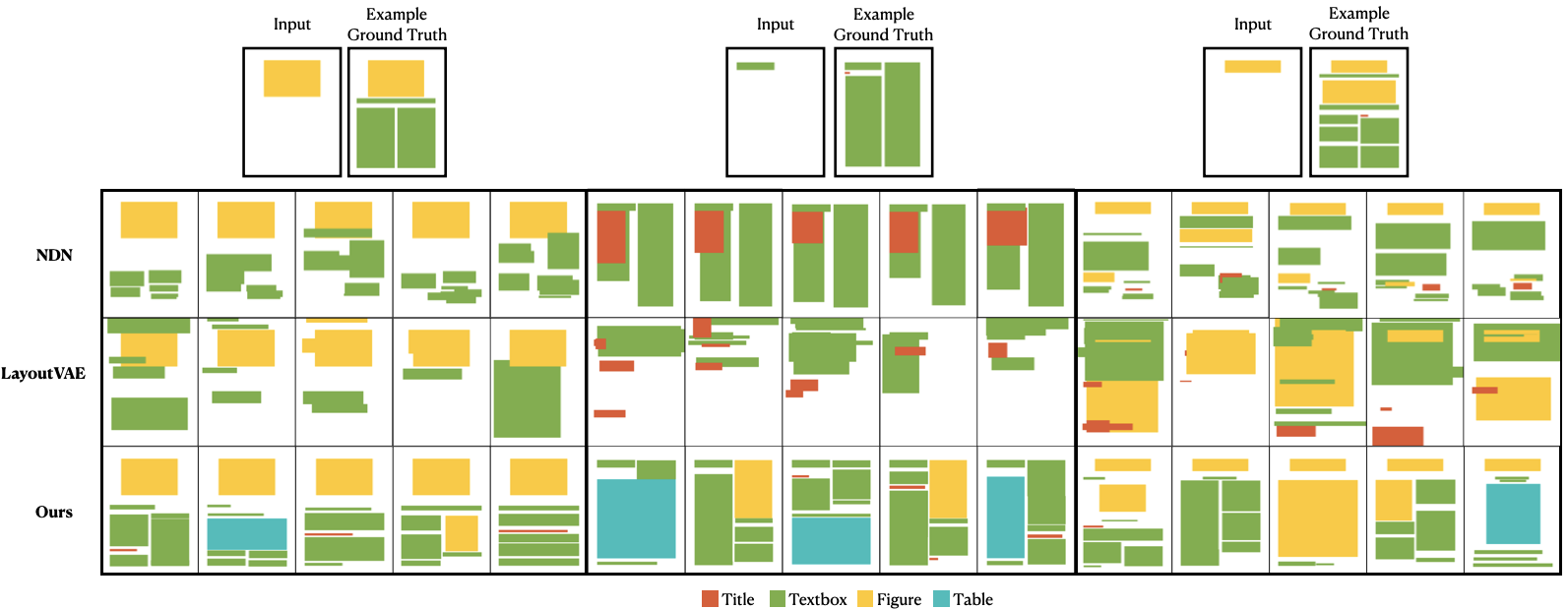}
  \caption{Each model generates multiple outputs using the same input to demonstrate diversity and consistency. 
  For NDN and LayoutVAE, the layouts shown are best 5 from 100 samples. 
  In contrast, LayoutMCL samples are the first 5 from 5.}
  \Description{Comparison of stability and diversity of outputs}
   \label{fig::diversitycomparison}
\end{figure*}

\subsection{Quantitative Metrics for Layouts}
We evaluate each model's ability to generate realistic layouts using metrics proposed in Section~\ref{sec:setup}.
In this experiment, LayoutVAE and LayoutMCL generates layouts by predicting categories and bounding boxes while NDN only predicts bounding boxes.
NDN is provided all object categories and all graph constraints prior to testing. 
To help the models generate samples within similar neighborhoods, we provide a single starting object for each layout. 
Each model generates 100 samples for each dataset and quantitatively compared in Table~\ref{tab:table1}.
Alignment and Fake Positive is calculated for a held-out test dataset and shown on the right-most column.

\begin{table}[h]
  \caption{Layout Generation Performance}
  \label{tab:table1}
  \begin{tabular}{lcccc}
    \toprule
     &LayoutVAE&NDN&Ours&Test\\
    \midrule
    \textbf{Mobile App} \\
    Alignment&0.01829&0.01767&\textbf{0.01477}&0.00659  \\
    FID&182.66&55.10&\textbf{8.15}&\\
    Fake Positive (\%) &93&54&\textbf{12}&5\\
    \midrule
    \textbf{Magazine} \\
    Alignment&0.04563&\textbf{0.02355}&0.029120&0.00926\\
    FID&1885.44&336.87&\textbf{26.18}&-\\
    Fake Positive (\%) &100&99&\textbf{48}&9\\
    \midrule
    \textbf{Document} \\
    Alignment&0.01434&0.01373&\textbf{0.00881}&0.00102\\
    FID&54.33&76.46&\textbf{9.13}&-\\
    Fake Positive (\%) &75&94&\textbf{23}&11\\
  \bottomrule
\end{tabular}
\end{table}

In summary, LayoutMCL outperforms both models across proposed metrics and datasets, with the exception of alignment on the Magazine dataset. 
For the mobile app and document dataset, LayoutMCL achieves the closest Alignment score to the held-out group and exceeds the next closest model by by 16-35\%. 
LayoutMCL achieves the best FID score which outperforms the next closest benchmark model by 83-98\% across all datasets.
Note that the FID score varies depending on the base model, checkpoints and composition of the held-out dataset.
The Mobile App FID scores are similar to prior reported \cite{lee2020neural}, however the Magazine FID scores are relatively higher here in comparison. 
Unfortunately, without open access to prior FID models, this discrepancy cannot be investigated.


\subsection{Diversity and Consistency Test}

The key feature of LayoutMCL is diversity and consistency of outputs.
Our experience with existing layout models is that they are able to generate visually appealing layouts, however, with low probability or consistency.
Consistency and diversity of results is critical in fully automated contexts such as decoy multimedia deployment in cybersecurity deception.

We create an experiment where every model is given the same input (partial layout) from which it repeatedly generates complete layouts successively.
Since all models achieved relatively low FID score for documents, we use this dataset as a benchmark.
Five samples for every model using 3 partial layouts are shown in Figure~\ref{fig::diversitycomparison}.

For NDN and LayoutVAE, the layouts shown here were selected as the best 5 from 100 generated. 
In contrast, LayoutMCL samples were the first 5 from 5 generated.
This demonstrates the significant advantage in reliability and consistency between the generative models. 
This can be attributable to the fact that NDN and LayoutVAE use a single distribution for bounding box prediction. 
These results show that quantitative measures do not translate directly into \textbf{consistent} visually appealing layouts. 

Figure~\ref{fig::diversitycomparison} also demonstrates the difference in sample diversity.
Every sample from LayoutMCL is a plausible document layout, each with a different composition of object types and sizes.
LayoutVAE fails to generate double columns as seen in the middle column and does not show diversity of categorical composition. 
Each group of samples are highly similar in their object type and count.
NDN also fails to produce diverse results as all layouts form similar patterns with the same input.

\subsection{Recommendations with Hard and Soft Constraints}
\begin{figure}[h]
  \centering
  \includegraphics[width=\linewidth]{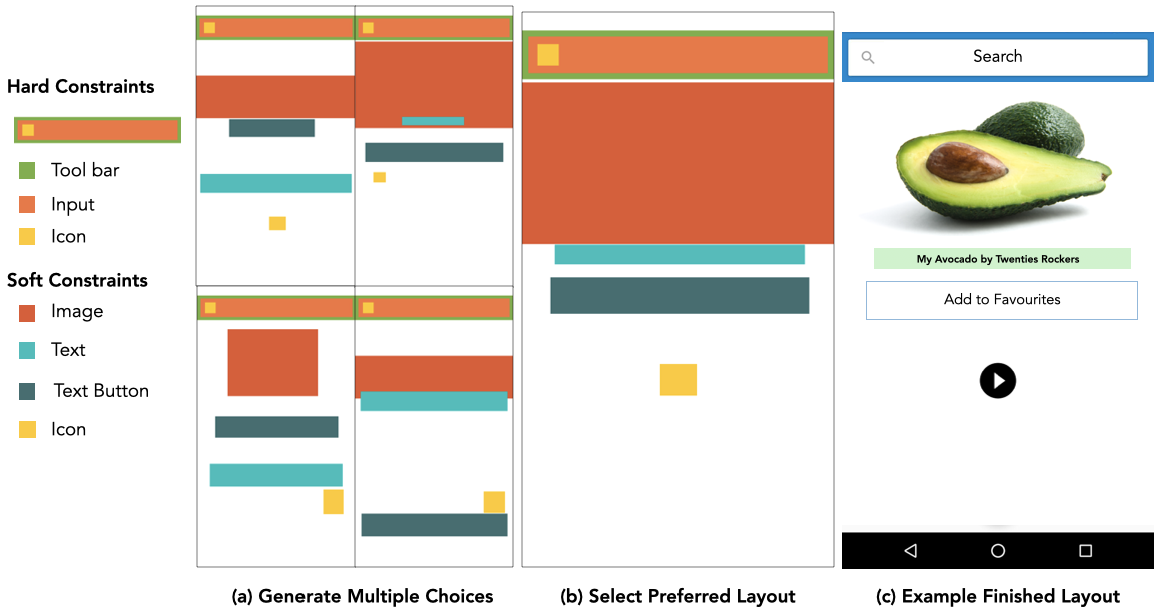}
  \caption{This visualisation demonstrates multi-choice layout recommendations with hard and soft constraints. This approach provides users with a range of potential options from which they can select from and create a final layout. Note that pictures in (c) are not generated by LayoutMCL. }
  \Description{The potential use case of LayoutMCL}
   \label{fig::usecase}
\end{figure}
In this section, we demonstrate the flexibility and benefits of our multi-choice framework through a simple use-case that combines both hard and soft constraints.
Hard constraints are preset object categories and locations that remain fixed during multi-hypothesis prediction.
This is useful where the designer wishes to keep elements of continuity between screens, such as a corporate logo or search bar.
Soft constraints are considered as a sequence of object properties that a designer wishes to obtain recommendations for in a final layout. 
This could include the object category or size but not location.
Within the framework, hard constraints are set as the prior sequence of objects and soft constraints are force fed into the multi-choice predictors at each step. 

A use-case for using hard and soft constraints within the framework is seen in Figure~\ref{fig::usecase}.
Here the user is provided multiple layout recommendations based on the constraints shown on the left. 
These soft and hard constraints can also be created in the context of layout rearrangement, where the user already has a layout but would like more options. 
Our multi-choice framework is beneficial for users because they are able to decide from a wide range of visual appealing layouts using \textbf{visual cues} as evident from the figure.
In comparison to prior methods, we find that this approach is user-friendly and intuitive in a real world application.









%% file: src/5_conclusion.tex
\section{Conclusion\label{cha:conclusion}}

In this paper we presented LayoutMCL, a layout generative model that contributes towards research in automated visual design.
We demonstrate that our auto-regressive architecture advances the existing state-of-art in layout generation by producing significantly more consistent, realistic and diverse samples across several metrics. 
These advancements are a result of combining several ideas from multi-choice learning and mixture density networks.
The core ideas demonstrated, multi-hypothesis prediction and winner-takes-all loss, can robustly learn diverse user preferences using multiple predictors. 
As a result, we demonstrate that reversing this process allows us to generate diverse layouts consistently.

For future work, we aim to study whether this framework can be applied in other potential domains such as procedural content generation for games, reinforcement learning or AR/VR design recommendations.




%% file: src/6_acks.tex
\begin{acks}
The authors would like to thank the Commonwealth of Australia and the Cybersecurity Cooperative Research Centre for their support.
\end{acks}